\documentclass{article}

\PassOptionsToPackage{numbers, compress, sort}{natbib}

\usepackage[preprint]{style_arxiv/neurips_2025}

\usepackage[utf8]{inputenc} 
\usepackage[T1]{fontenc}    
\usepackage{hyperref}       
\usepackage{url}            
\usepackage{booktabs}       
\usepackage{amsfonts}       
\usepackage{nicefrac}       
\usepackage{microtype}      
\usepackage{xcolor}         

\newif\ifarxiv
\arxivtrue

\usepackage{shortcuts}

\title{AnCoder: Anchored Code Generation via Discrete Diffusion Models}

\author{%
  Anton Xue\thanks{Equal contribution.} \quad
  Litu Rout\footnotemark[1] \quad
  Constantine Caramanis \quad
  Sanjay Shakkottai \\
  The University of Texas at Austin \\
  \texttt{\{anton.xue, litu.rout, constantine, sanjay.shakkottai\}@utexas.edu}
}

\begin{document}

\maketitle

\begin{abstract}

Diffusion language models offer a compelling alternative to autoregressive code generation, enabling global planning and iterative refinement of complex program logic.
However, existing approaches fail to respect the rigid structure of programming languages and, as a result, often produce broken programs that fail to execute.
To address this, we introduce AnchorTree, a framework that explicitly anchors the diffusion process using structured, hierarchical priors native to code.
Specifically, AnchorTree uses the abstract syntax tree to prioritize resolving syntactically and semantically salient tokens, such as keywords (e.g., \texttt{if}, \texttt{while}) and identifiers (e.g., variable names), thereby establishing a structural scaffold that guides the remaining generation.
We validate this framework via AnCoder, a family of models showing that structurally anchored diffusion offers a parameter-efficient path to high-quality code generation.





\end{abstract}

\label{sec:introduction}

\ifarxiv
    \begin{figure}[h]
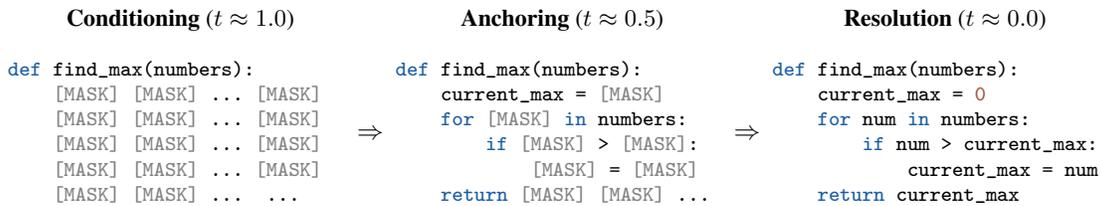

\centering

\centering

\begin{minipage}[t]{0.31\linewidth}
    \centering
    {\small\textbf{\textrm{Conditioning}} ($t \approx 1.0$)} \\
    {\fontsize{8.5}{9.5}\ttfamily
    \vspace{4pt} 
    \begin{flushleft}
        \kwd{def} find\_max(numbers):\\
        \ \ \ \ \mask\ \mask\ \dots\ \mask\ \\
        \ \ \ \ \mask\ \mask\ \dots\ \mask\ \\
        \ \ \ \ \mask\ \mask\ \dots\ \mask\ \\
        \ \ \ \ \mask\ \mask\ \dots\ \mask\ \\
        \ \ \ \ \mask\ \mask\ \dots\ \ \dots\
    \end{flushleft}
    }
\end{minipage}%
\hfill \raisebox{-1.5cm}{$\Rightarrow$} \hfill 
\begin{minipage}[t]{0.30\linewidth}
    \centering
    {\small\textbf{\textrm{Anchoring}} ($t \approx 0.5$)} \\
    {\fontsize{8.5}{9.5}\ttfamily
    \vspace{4pt}
    \begin{flushleft}
        \kwd{def} find\_max(numbers):\\
        \ \ \ \ current\_max = \mask \\
        \ \ \ \ \kwd{for} \mask\ \kwd{in} numbers:\\
        \ \ \ \ \ \ \ \ \kwd{if} \mask\ > \mask:\\
        \ \ \ \ \ \ \ \ \ \ \ \ \mask\ = \mask\\
        \ \ \ \ \kwd{return} \mask\ \mask\ \dots
    \end{flushleft}
    }
\end{minipage}%
\hfill \raisebox{-1.5cm}{$\Rightarrow$} \hfill
\begin{minipage}[t]{0.31\linewidth}
    \centering
    {\small\textbf{\textrm{Resolution}} ($t \approx 0.0$)} \\
    {\fontsize{8.5}{9.5}\ttfamily
    \vspace{4pt}
    \begin{flushleft}
        \kwd{def} find\_max(numbers):\\
        \ \ \ \ current\_max = \num{0}\\
        \ \ \ \ \kwd{for} num \kwd{in} numbers:\\
        \ \ \ \ \ \ \ \ \kwd{if} num > current\_max:\\
        \ \ \ \ \ \ \ \ \ \ \ \ current\_max = num\\
        \ \ \ \ \kwd{return} current\_max
    \end{flushleft}
    }
\end{minipage}

\vspace{4pt}

\caption{\textbf{AnCoder: Anchored Code Generation.}
Generation begins with \textit{conditioning}, where the input prompt is mask-padded.
During the \textit{anchoring} phase, syntactically and semantically salient tokens such as keywords and identifiers are unmasked according to their syntactic hierarchy.
Given these anchors, the model can then better unmask the remaining tokens in the final \textit{resolution} phase.
Denoising in this order yields code with higher functional correctness than diffusion language model baselines.
}
\label{fig:teaser}
\end{figure}
\else
\fi

\section{Introduction}

LLM-powered coding agents are now widely used in software engineering on tasks from simple scripting to architecting complex infrastructure~\citep{ahmad2021unified,alphacode,bavarian2022efficient,claude,gemini,gpt,deepseek,qwen3}.
However, most implementations rely on autoregressive (AR) generation, which decodes tokens sequentially from left to right.
By imposing sequential generation on inherently non-sequential structures such as syntax and control flow, AR models incur a structural mismatch that limits the global planning required for high-quality code~\citep{incoder}.
Moreover, this AR process stands in contrast with how humans reason, as it neglects the ``coarse-to-fine'' strategies essential for managing large codebases~\citep{heinonen2023synthesizing}.
Consequently, LLM-generated code frequently exhibits structural idiosyncrasies and ``code smells'' that create cognitive friction for human maintainers~\citep{siddiq2024quality,abbassi2025unveiling,santa2025llm}.

A key source of these frictions is that code is fundamentally distinct from natural-language sequences~\citep{hindle2016naturalness,casalnuovo2019studying}.
Syntactically, parentheses, brackets, and indentation must be balanced across multiple lines;
semantically, variables must be well-scoped, and execution may branch, loop, or occur out-of-order with respect to function definitions~\citep{muller2014pystruct,incoder}.
Whereas natural language is typically robust to minor grammatical and wording mistakes, any violation in the program logic may easily lead to code that fails to run~\citep{lam2025codecrash,dong2025survey,sharifloo2025llms}.
These differences motivate the need to explore alternatives to sequential code generation, especially ones that can reason about global structure and self-correct potential mistakes~\citep{olausson2023self}.



Diffusion language models (DLMs) provide one such alternative and have recently been explored for code generation~\citep{gong2025diffucoder,xie2025dream,chen2025coda,mercurydllm2025,li2025beyond}.
By framing generation as an iterative denoising process from a noisy token sequence, DLMs first sketch high-level structures and constraints, then gradually refine them with low-level implementation details.
This coarse-to-fine generation not only mirrors the cognitive strategies of human programmers~\citep{heinonen2023synthesizing}, but also unlocks superior length extrapolation, allowing DLMs to maintain coherence on long sequences where autoregressive models degrade~\citep{li2025beyond}.




Nevertheless, even DLMs trained for code can struggle to produce correct programs.
Like early versions of AR coding models, DLMs can just as easily generate programs that crash~\citep{li2025beyond}.
To mitigate these issues, a growing body of work has proposed test-time adaptations for quality generation, such as constrained decoding methods to ensure syntactically valid code~\citep{suresh2025dingo,mundler2025constrained}.
While effective and complementary to our work, these methods are inherently post-hoc: they apply corrections during generation, rather than improving the model's understanding of code structures through training.


Our key observation is that code exhibits inherent hierarchical structure at every stage of its lifecycle.
During parsing, compilers first lift token sequences into an abstract syntax tree (AST) to resolve grammatical ambiguity~\citep{aho2007compilers}.
This hierarchical nature re-emerges in intermediate analysis, where code optimizations rely on dominator trees, distinct from syntax, to reason about control flow and loop nesting~\citep{cooper2022engineering}.
Similarly, machine code generation is frequently modeled as a tree-covering problem in which expression subtrees are mapped to instruction sequences~\citep{lattner2004llvm}.
Finally, hierarchy governs execution itself: the runtime call stack organizes function calls to ensure that variable lifetimes remain strictly nested~\citep{scott2000programming}.
The prevalence of hierarchy in code suggests that successful diffusion language modeling requires not only an understanding of token-level distributions but also hierarchical reasoning through multiple levels of abstraction.


A recent approach presents \textit{anchoring} as an effective framework to encode hierarchical structures in DLM training and generation~\citep{rout2025anchored,rout2025test}.
The main idea is that a DLM should first unmask the informative output tokens (e.g., nouns, verbs), which then ``anchors'' the generation of the surrounding details.
Notably, anchored DLMs achieve significant performance gains on language modeling benchmarks compared to their non-anchored counterparts, often with only a fraction of the parameters.




However, existing approaches implement \textit{hard anchoring} schemes that cannot capture code-specific hierarchies.
In hard anchoring, a token is either anchored or not, and all anchored tokens are treated with equal weight.
While this establishes a hierarchy between anchored and unanchored tokens, there is no further distinction among the anchors.
Consequently, hard anchoring would not distinguish between anchors in top-level function definitions and deeply nested for loops, nor would it distinguish between anchors in the two branches of an if-else statement.





To address the limitations of hard anchoring, we introduce \textbf{AnchorTree}, a \textit{soft anchoring} framework tailored to code-specific hierarchies.
In AnchorTree, we first select keywords (e.g., \texttt{if}, \texttt{for}) and identifiers (e.g., variable names) as anchor tokens because they specify syntactic structure and semantic intent~\citep{hindle2016naturalness}.
Then, we assign each anchor a weight based on its position in the AST, with tokens closer to the root node receiving more weight.
This soft anchoring scheme encourages syntactically and semantically salient tokens to be denoised in the order of their AST-based syntactic hierarchy, which we highlight in~\cref{fig:teaser}.


Our contributions are summarized as follows:
\textbf{(1)} We introduce AnchorTree, a soft anchoring framework that enables DLMs to better learn the syntactic and semantic hierarchical structures of code.
\textbf{(2)} Using AnchorTree, we train a family of anchored diffusion models that we call \textbf{AnCoder}, and show that it outperforms DLM baselines on standard coding benchmarks such as HumanEval and MBPP.
\textbf{(3)} To the best of our knowledge, we are the first to demonstrate that soft, hierarchical anchoring schemes can significantly improve code generation quality of diffusion language models.

\section{Related Work}
\label{sec:related}
\textbf{Diffusion Language Models}~\citep{li2022diffusion,llada} generate token sequences by learning to reverse a gradual noising process, typically consisting of random token masking~\citep{lou2023discrete,shi2024simplified,ou2024your,sahoo2024simple} or flipping~\citep{austin2021structured,lou2023discrete}.
In practice, DLM generation is implemented by starting with mask-padded input sequence and iteratively unmasking (or flipping) tokens until a final coherent output remains.
This process exhibits the ability to self-correct generation mistakes~\citep{remdm} and to improve global coherence, particularly for structured outputs~\citep{li2025beyond}.
However, DLMs often underperform their AR counterparts in generation quality~\citep{llada}.
A key reason for this is because informative tokens (e.g., nouns, verbs) are often revealed too late in the denoising process~\citep{rout2025anchored}, which deprives earlier generation stages of vital conditioning information.

\textbf{Anchored Diffusion Models.}
To address the problem of missing informative tokens, \citet{rout2025anchored} introduce \emph{anchored} diffusion language models that explicitly prioritize the denoising of such tokens. 
By revealing these ``anchor tokens'' early in the denoising process, the model conditions subsequent denoising steps on a more informative context. 
Anchoring has been shown to be effective across several domains, including natural language generation and mathematical reasoning in autoregressive models~\citep{rout2025anchored}, and can also be implemented with minimal architectural changes via test-time adaptation~\citep{rout2025test}.
However, existing methods label tokens as hard anchors, which is not necessarily effective for code generation.

\textbf{Diffusion Models for Code Generation.}
A growing body of work explores the use of DLMs for code generation, including DiffuCoder~\citep{gong2025diffucoder}, Dream~\citep{xie2025dream}, CoDA~\citep{chen2025coda}, Mercury~\citep{mercurydllm2025}, and Gemini Diffusion~\citep{geminidiffusion}. 
These approaches primarily adapt diffusion language models to code datasets, focusing on architectural choices, training objectives, or decoding strategies.
While they demonstrate the feasibility of diffusion-based code generation~\citep{li2025beyond}, they do not explicitly encode program structure during training, and as a result, remain vulnerable to syntactic validity and semantic errors~\citep{suresh2025dingo,mundler2025constrained}.

\textbf{Structure-Aware Diffusion for Code.}
In standard masked diffusion, each position of a clean sequence is independently masked and a denoiser model is trained to reconstruct the original tokens.
TreeDiff~\citep{zeng2025treediff} modifies this by introducing a forward noising schedule that is not position-wise independent and also depends on AST depth.
However, TreeDiff leaves the denoising process unchanged, using a standard single-stage decoder.
In contrast, we explicitly modify the denoising to use a two-stage architecture: we first use the \textit{anchor network} to prioritize the unmasking of informative tokens in the order of their syntactic hierarchy, which then acts as informative anchors for the \textit{denoiser network} to unmask the remaining tokens.

\section{Preliminaries}
\label{sec:background}

\textbf{Notations.}
Let \(\mcal{V} = \{1, \ldots, K - 1, K\}\) denote the set of tokens in a vocabulary where \(K\) is a special ``mask'' token.
Given an \(L\)-length token sequence \(x = (x^1, \ldots, x^L) \in \mcal{V}^L\), we equivalently represent it as a sequence of one-hot vectors \(\mbf{x} = (\mbf{x}^1, \ldots, \mbf{x}^L)\), where each \(\mbf{x}^l \in \{0,1\}^{K}\) is hot only at position \(x^l\).
For the special mask token, let \(\mbf{m} = (0, \ldots, 0, 1)^\top\) denote its one-hot encoding.
We assume that the training data \(\mbf{x}\) is sampled from an unknown distribution \(q(\cdot)\) supported on \(\mcal{V}^L\).

\subsection{Discrete Diffusion Language Models}
\label{sec:prelim-dlm}
Discrete diffusion language models~\citep{sohl2015deep,austin2021structured,lou2023discrete,sahoo2024simple,shi2024simplified} are based on a forward process that gradually noises a clean token sequence \(\mbf{x} = (\mbf{x}^1, \ldots, \mbf{x}^L)\) into a fully masked sequence as follows:
\begin{equation} \label{eqn:forward_process_overview}
    (\mbf{x}^1, \ldots, \mbf{x}^L)
    \longrightarrow
    (\mbf{z}_t ^1, \ldots, \mbf{z}_t ^L)
    \longrightarrow (\mbf{m}, \ldots, \mbf{m}),
\end{equation}
where time flows from \(t = 0\) (all clean) to \(t = 1\) (fully masked) and each \(\mbf{z}_t ^l\) is a latent, one-hot vector that equals either \(\mbf{x}^l\) or \(\mbf{m}\) at random.
This forward-noising process happens independently at each position \(l\) and is defined by the conditional distribution
\begin{equation}
    q(\mbf{z}_t | \mbf{x}) = \prod_{l = 1}^{L} q(\mbf{z}_t ^l | \mbf{x}),
    \quad
    q(\mbf{z}_t ^l | \mbf{x})
    = \mrm{Cat}(\mbf{z}_t ^l; \alpha_t \mbf{x}^l + (1 - \alpha_t) \mbf{m}),
\end{equation}
and \(\alpha_t \in [0,1]\) is a monotonically decreasing noise schedule satisfying \(\alpha_0 = 1\) and \(\alpha_1 = 0\).
In practice, we discretize the forward process into \(T\) time steps, where \(t(i) = \tfrac{i}{T}\) and \(s(i) = \tfrac{i - 1}{T}\) for indicies \(i \in \{1, \ldots, T\}\).
When \(i\) is not specified, let \(s, t\) denote any two consecutive time steps.

The objective of generative diffusion modeling is to learn the reverse process, i.e., reversing the flow of time to denoise a fully masked sequence into a clean one.
To do this, observe that the forward conditional at each position is a Markov process with transition probability:
\begin{equation} \label{eqn:forward_markov_transition}
    q(\mbf{z}_t ^l | \mbf{z}_s ^l)
    = \mrm{Cat}\parens*{\mbf{z}_t ^l; \tfrac{\alpha_t}{\alpha_s} \mbf{z}_s ^l + (1 - \tfrac{\alpha_t}{\alpha_s}) \mbf{m}},
\end{equation}
where \(\mbf{z}_0 ^l = \mbf{x} ^l\).
Intuitively, this states that once a position is masked, it remains masked for the rest of the forward process.
We aim to reverse this forward Markov transition; \textit{suppose} the clean target \(\mbf{x} \sim q(\cdot)\) is known, the optimal reverse transition then has the form
\begin{equation} \label{eqn:reverse_posterior}
    q(\mbf{z}_s ^l | \mbf{z}_t ^l, \mbf{x}^l) =
    \begin{dcases}
        \mrm{Cat} \parens*{\mbf{z}_s^l ; \mbf{z}_t^l},
            & \mbf{z}_t^l \neq \mbf{m} \\
        \mrm{Cat} \parens*{\mbf{z}_s^l; \tfrac{\alpha_s - \alpha_t}{1 - \alpha_t} \mbf{x}^l + \tfrac{1 - \alpha_s}{1 - \alpha_t} \mbf{m}},
            & \mbf{z}_t^l = \mbf{m}.
    \end{dcases}
\end{equation}
That is, unmasked positions remain unmasked, while masked positions are probabilistically flipped to \(\mbf{x}^l\).

However, the clean target sequence \(\mbf{x}\) is not known during generation, i.e., we do not have a sample \(\mbf{x} \sim q(\cdot)\), so we cannot directly implement~\cref{eqn:reverse_posterior}.
Instead, the idea is to use a \(\theta\)-parametrized model that predicts what \(\mbf{x}\) \textit{should be} given the current iterate \(\mbf{z}_t\).
We express this prediction as
\begin{equation}
    \mbf{x}_\theta (\mbf{z}_t) = (\mbf{x}_\theta ^1 (\mbf{z}_t), \ldots, \mbf{x}_\theta ^L (\mbf{z}_t)),
\end{equation}
where each \(\mbf{x}_\theta ^l (\mbf{z}_t)\) is a probability vector over \(\mcal{V}\) from which we have the identity \(p_\theta (\mbf{x}^l | \mbf{z}_t) = \angles{\mbf{x}_\theta ^l (\mbf{z}_t), \mbf{x}^l}\).
We further employ two constraints: \textit{zero-masking}, where each \(\mbf{x}_\theta ^l (\mbf{z}_t)\) has zero mass on the mask token, and \textit{carry-over unmasking}, where unmasked \(\mbf{x}_\theta ^l (\mbf{z}_t) = \mbf{z}_t ^l\) whenever \(\mbf{z}_t ^l \neq \mbf{m}\).
Using these, we then define the \(\theta\)-parametrized backward-denoising process as follows:
\begin{equation}
p_\theta (\mbf{z}_s | \mbf{z}_t) = \prod_{l = 1}^{L} p_\theta (\mbf{z}_s ^l | \mbf{z}_t),
\quad
p_\theta (\mbf{z}_s ^l | \mbf{z}_t) =
    \begin{dcases}
        \mrm{Cat}\parens*{\mbf{z}_s ^l; \mbf{z}_t ^l},
            & \mbf{z}_t ^l \neq \mbf{m} \\
        \mrm{Cat}\parens*{\mbf{z}_s ^l; \tfrac{\alpha_s - \alpha_t}{1 - \alpha_t} \mbf{x}_\theta ^l (\mbf{z}_t) + \tfrac{1 - \alpha_s}{1 - \alpha_t} \mbf{m}},
            &\mbf{z}_t ^l = \mbf{m}.
    \end{dcases}
\end{equation}
The \(\theta\)-parametrized denoising process is trained using the standard Negative ELBO (NELBO) framework for maximum likelihood estimation.
For the diffusion scheme described here, the loss is equivalent to
\begin{equation}
\label{eq:nelbo}
    \mcal{L}_{\mrm{NELBO}} (\mbf{x}; \theta) =
    \sum_{i = 1}^{T} \expval{\mbf{z} \sim q(\cdot | \mbf{x})} \bigg[
        \lambda_{i}
        \sum_{l = 1}^{L}
        \log \angles{\mbf{x}_\theta ^l (\mbf{z}_{t(i)}), \mbf{x}^l}
        \bigg],
        \quad
        \lambda_i = \frac{\alpha_{t(i)} - \alpha_{s(i)}}{1 - \alpha_{t(i)}},
\end{equation}
where \(\lambda_i\) is the weight at time index \(i\) due to the noise schedule.

\subsection{Anchored Diffusion Language Models}
\label{sec:anchored-dlm}

\ifarxiv
\begin{figure}[t]
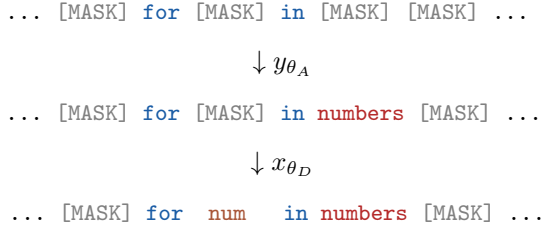


\centering

\begin{minipage}[t]{0.5\linewidth}
    \centering
    \small\ttfamily
    \begin{flushleft}
        \dots\ \mask\ \kwd{for} \mask\ \kwd{in} \mask\ \mask\ \dots
    \end{flushleft}
\end{minipage}
\vspace{8pt}\\

\(\downarrow y_{\theta_A}\) \vspace{8pt} \\

\begin{minipage}[t]{0.5\linewidth}
    \centering
    \small\ttfamily
    \begin{flushleft}
        \dots\ \mask\ \kwd{for} \mask\ \kwd{in} \anc{numbers}\ \mask\ \dots
    \end{flushleft}
\end{minipage}
\vspace{8pt}

\(\downarrow x_{\theta_D}\) \vspace{8pt}

\begin{minipage}[t]{0.5\linewidth}
    \centering
    \small\ttfamily
    \begin{flushleft}
        \dots\ \mask\ \kwd{for} \ \num{num}\ \ \ \kwd{in} \anc{numbers}\ \mask\ \dots
    \end{flushleft}
\end{minipage}

\vspace{4pt}

\caption{\textbf{ADLM forward pass.}
Given a partially masked \(z_t\) (top), the anchor network \(y_{\theta_A}\) focuses on unmasking the anchor token {\color{codered}\texttt{numbers}} (middle), which helps the denoiser network \(x_{\theta_D}\) unmask the remaining tokens, such as {\color{codeorange}\texttt{num}} (bottom).
Anchor tokens are labeled prior to training (see~\cref{sec:experiments}).
}
\label{fig:anchor_framework_overview}
\end{figure}
\fi

A limitation of standard DLMs is that tokens critical for generation may only be unmasked late in the denoising process.
When such tokens remain masked for most of the denoising trajectory, the model must generate dependent content without sufficient conditioning information, which can degrade generation quality.
To address this issue, \citet{rout2025anchored} introduce \emph{anchored} diffusion language models, which explicitly condition generation on a subset of important tokens, referred to as \emph{anchor tokens}.
The core intuition is that the presence of salient tokens simplifies the generation of the remaining tokens, see~\cref{fig:anchor_framework_overview}.


\textbf{Anchorized Diffusion Language Models (ADLM).}
Anchored diffusion decomposes a single denoising prediction into two stages, wherein the standard prediction network $\mbf{x}_\theta(\mbf{z}_t)$ is rewritten as the composition
\begin{equation}
    \mbf{x}_\theta(\mbf{z}_t)
    =
    \mbf{x}_{\theta_D}\bigl(\mbf{y}_{\theta_A}(\mbf{z}_t)\bigr),
\end{equation}
where \(\theta = \{\theta_A, \theta_D\}\) is a partition of parameters for the \textit{anchor network} \(\mbf{y}_{\theta_A}\) and \textit{denoiser network} \(\mbf{x}_{\theta_D}\).
Both networks output probability vectors over the same vocabulary \(\mcal{V}\), and we refer to this parametrization as an anchored diffusion language model.

\textbf{Anchored Training Objective.}
Training proceeds by supervising both the final denoising prediction and the intermediate anchor prediction.
This yields the Anchored Negative ELBO (ANELBO) objective:
\begin{equation}
\label{eq:anelbo}
\mcal{L}_{\mrm{ANELBO}} (\mbf{x}; \theta_A, \theta_D) =
    \sum_{i = 1}^{T} \expval{\mbf{z} \sim q(\cdot | \mbf{x})} \bracks*{
    \lambda_i \sum_{l = 1}^{L}
    \underbrace{\log \angles{\mbf{x}_{\theta} ^l (\mbf{z}_{t(i)}), \mbf{x}^l}}_{\text{NELBO term}} + \underbrace{\mu \log \angles{\mbf{y}_{\theta_A} ^l (\mbf{z}_{t(i)}), \mbf{y}^l}}_{\text{Anchor term}}
    },
\end{equation}
where $\mu > 0$ controls the strength of anchor supervision and $\mbf{y}^l$ denotes the anchor label at position $l$.
This applies auxiliary supervision to an intermediate prediction to encourage the recovery of anchor tokens earlier during denoising.

\textbf{Anchor Specification.}
Anchored diffusion models require anchor supervision to be specified prior to training via a preprocessing step that determines which tokens receive anchor labels.
Prior work~\citep{rout2025anchored} defines anchors using task-dependent heuristics such as token frequency, digit extraction for mathematical reasoning, or parts-of-speech tagging for natural language tasks.
These approaches induce \emph{hard anchoring}, where each token is either designated as an anchor or treated as a non-anchor.

However, anchoring need not be so binary.
In general, tokens may exhibit varying degrees of structural importance, suggesting a \emph{soft anchoring} formulation in which tokens receive graded anchor supervision.
In this work, we define anchor specification using structural priors native to code.
We next present such a framework in~\cref{sec:anchortree}.


\section{AnCoder: Anchored Code Generation via Abstract Syntax Trees}
\label{sec:anchortree}

Our core method is to anchor DLM generation with hierarchical structures native to code, in particular, the abstract syntax tree (AST).
In the following, we provide an overview of the AST and its potential for anchoring.
Then, we present an experiment to highlight that the hierarchical information provided in the AST is, in fact, useful for the denoising process.
We show how to translate these insights into an effective soft anchoring strategy, \textbf{AnchorTree}, which we leverage to train the \textbf{AnCoder} family of DLMs.


\begin{figure}[!t]

\centering




\ifarxiv
    \includegraphics[width=0.90\linewidth]{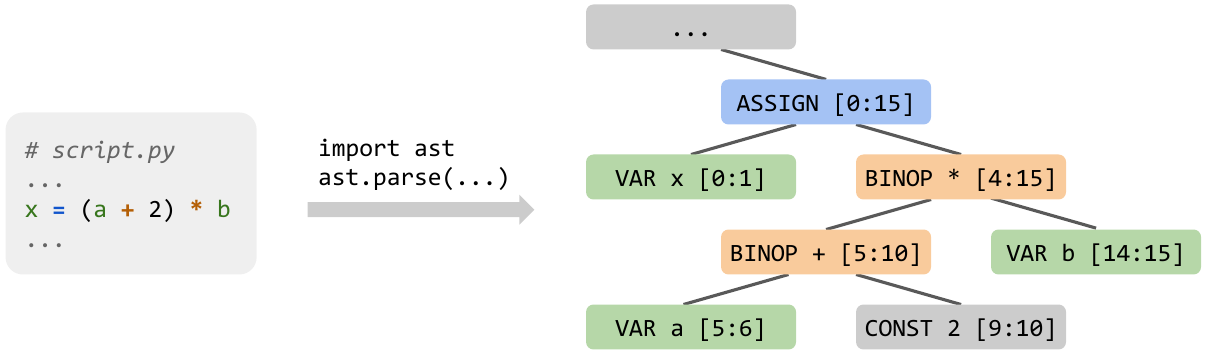}
\fi


\caption{
\textbf{The syntactic hierarchy of code.}
A sequence of source code tokens (left) is parsed into an AST (right).
Here, a single assignment statement forms a subtree nested within the broader program context (denoted by \texttt{...}).
AST nodes are labeled with the type, data, and relative character spans (e.g., \texttt{[5:10]}) that map high-level syntactic constructs onto their sequential token positions.
}
\label{fig:ast_example}
\end{figure}

\subsection{The Abstract Syntax Tree}
\label{sec:ast_background}

The AST is a tree-structured representation of program syntax used in program compilation and analysis~\footnote{Python AST specification: \url{https://docs.python.org/3/library/ast.html}}~\citep{aho2007compilers}.
\cref{fig:ast_example} shows an example, where each AST node is labeled with its syntactic type (e.g., assignment, binary operation), relevant data (e.g., variable names or literal values), and a relative character span indicating the start and end positions of its associated region of code.
Specifically, \cref{fig:ast_example} shows the abstract syntax (sub)-tree of the statement ``\texttt{x = (a + 2) * b}'' embedded within a broader Python script.
Here, the character span \texttt{[4:15]} corresponds to the sub-expression \texttt{(a + 2) * b}, while \texttt{[5:10]} corresponds to the inner binary operation \texttt{a + 2}, where note the presence of whitespace characters.







ASTs have two properties useful for anchored diffusion.
First, AST nodes are grouped and labeled by syntactic constructs, such as variable assignments, function calls, and class definitions, which provides a rich design space for anchor selection.
For instance, one might wish to anchor specifically on variables within function signatures or return statements, which is easily achieved by standard tree-traversal algorithms over the AST.
Second, the AST defines a canonical syntactic hierarchy of source code in that all expressions and sub-expressions admit a natural partial ordering under span inclusions.
This hierarchy naturally aligns with diffusion's notion of coarse-to-fine denoising: roughly speaking, we would like to unmask tokens closer to the AST root (e.g., often function and class definitions) before those closer to the leaves (e.g., deeply nested if statements).


\textbf{AST-based Hierarchy on Tokens.}
To formalize the connection between syntax and denoising, we map the AST hierarchy onto the tokenized sequence $\mbf{x} = (\mbf{x}^1,\ldots,\mbf{x}^L)$.
First, we associate each position \(l\) with the deepest AST node whose character span intersects that of \(\mbf{x}^l\), and call this node as \(\msf{node}(l)\)~\footnote{
This accounts for potential character-level boundary misalignment between DLM tokenizers and AST parsers.
}.
We then define a partial order over token positions by declaring $l \succeq l'$ if either $\msf{node}(l)$ is a strict ancestor of $\msf{node}(l')$ in the AST, or $\msf{node}(l)=\msf{node}(l')$ and $l<l'$.
The latter case occurs, for example, when a tokenizer splits an identifier such as \texttt{quicksort} into multiple tokens.
Under this partial ordering, one may interpret \(l \succeq l'\) to mean that \(l\) is a syntactically coarser position than \(l'\).
In other words, positions higher in the AST correspond to syntactically coarser elements of the program, while positions deeper in the tree correspond to finer-grained details.

\begin{figure*}[t]

\centering

\begin{minipage}{0.35\linewidth}
    \centering
    \includegraphics[height=1.24in]{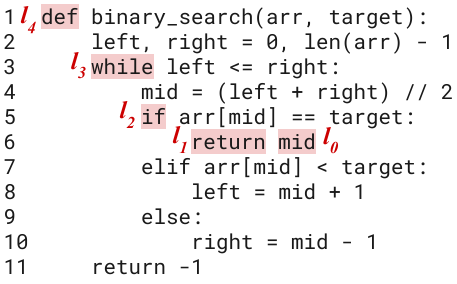}
\end{minipage}%
\hfill
\begin{minipage}{0.32\linewidth}
    \centering
    \includegraphics[height=1.24in]{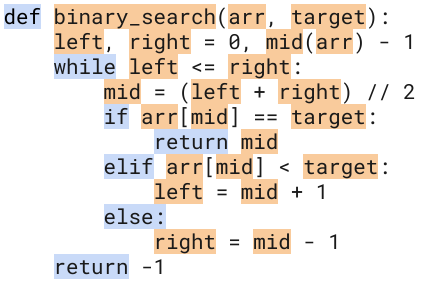}
\end{minipage}%
\hfill
\begin{minipage}{0.32\linewidth}
    \centering
    \includegraphics[height=1.24in]{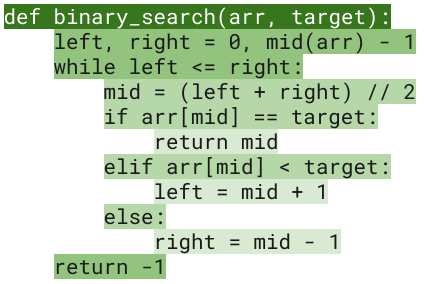}
\end{minipage}%

\vspace{4pt}

\caption{\textbf{AST-based ordering and the AnchorTree weight \(\mu(l) = \omega(l) \cdot \eta(l)\).}
(Left) Example of an ascending chain \(l_0 \preceq l_1 \preceq \cdots \preceq l_4\) starting at \texttt{mid} and ordered by the AST hierarchy.
(Middle) We take keywords (blue) and identifiers (orange) to be the anchors, where let \(\omega(l) = 1\).
(Right) Each position is weighted by the AST depth, where let \(\eta (l') \geq \eta(l)\) if \(l' \succeq l\) in the AST-based partial ordering.
}
\label{fig:anchor_strategies_gallery}
\end{figure*}

\begin{figure}[!t]

\centering


\ifarxiv
    \includegraphics[width=0.55\linewidth]{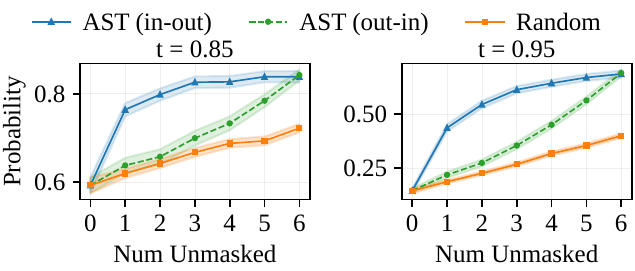}
\fi

\caption{
\textbf{Unmasking by AST ancestry improves performance.}
Cumulatively unmasking a target position \(l_0\)'s AST ancestors \(l_1, l_2, \ldots\) improves performance more than simply unmasking other tokens at random.
Moreover, performance improves the fastest when we reveal AST ancestors closest to the target first (in-out).
This trend holds at both moderate-noise (\(t = 0.85\)) and high-noise (\(t = 0.95\)) settings.
}
\label{fig:ast_ancestor_probs}
\end{figure}
\subsection{Is the AST Useful for Denoising?}
Suppose that we are given a partially masked sequence in which position \(l\) is masked, and that we are allowed to unmask another position as a hint for \(l\).
If syntactic ancestry is useful, then we should expect that unmasking an ancestor of \(l\) is preferable to unmasking a non-ancestor.

We test this hypothesis on a pretrained DLM and a code dataset~\footnote{
Using the pretrained 809M model (Stage 1 and 2) with 2000 test samples from the SFT dataset. See~\cref{sec:experiments}.
}.
Let \(\mbf{x}\) be a tokenized sequence of code and \(\mbf{z}_t\) be its noised version.
We additionally mask \(\mbf{z}_t\) at positions \(l_0, l_1, \ldots, l_k\), which we specifically pick to be a contiguous chain in the AST.
That is, \(l_0 \preceq l_1 \preceq \cdots \preceq l_k\) and there does not exist another \(l'\) such that \(l_i \preceq l' \preceq l_{i+1}\) for any \(i\).
This lets us isolate and test how unmasking the ancestors of \(l_0\) might improve the model prediction at \(l_0\).
As an example, using the binary search implementation in~\cref{fig:anchor_strategies_gallery} (left), where let \(l_0\) correspond to the \texttt{mid} token within the \texttt{return mid} line (line 6 in \cref{fig:anchor_strategies_gallery}).
One possible chain of AST ancestor tokens starting from \texttt{mid} (\(l_0\)) would thus be \texttt{return} (\(l_1\)), \texttt{if} (\(l_2\)), \texttt{while} (\(l_3\)), and \texttt{def} (\(l_4\)).

We consider two strategies for unmasking \(l_1, \ldots, l_k\), sorted by their AST-based proximity to \(l_0\).
In the \textit{in-out} order, we progressively unmask the closest AST ancestors to \(l_0\) first; in the \textit{out-in} ordering, we reverse this and unmask the furthest ancestors first.
In the example above: we would begin with \texttt{mid} (\(l_0\)), \texttt{return} (\(l_1\)), \texttt{if} (\(l_2\)), \texttt{while} (\(l_3\)), and \texttt{def} (\(l_4\)) all masked.
Then, for the in-out order, we first unmask \texttt{return} (\(l_1\)), then \texttt{if} (\(l_2\)), etc., and at each unmasking iteration querying the model's prediction at the masked position \(l_0\).
Conversely, for the out-in order, we first unmask \texttt{def} (\(l_4\)), then \texttt{while} (\(l_3\)), etc., similarly querying the model's prediction at \(l_0\) during each iteration.

We compare the in-out and out-in unmasking orders with a baseline in which a number of random positions from outside this chain are unmasked instead.
We show the results in \cref{fig:ast_ancestor_probs}, where we observe that unmasking ancestor positions significantly improves the prediction at the \(l_0\)-th position.
In particular, progressively unmasking the closest ancestors first (in-out) improves the probability of predicting the correct label at \(l_0\) the fastest.

\subsection{AST-based Hierarchical Soft Anchoring}
\label{sec:anchor-tree}

The main insight above is that revealing the ancestors of a masked position can improve prediction.
Moreover, the closer the revealed ancestor, the greater the performance improvement.
This suggests an anchored denoising scheme that gradually unmasks tokens based on the syntactic hierarchy, with nodes closest to the AST root unmasked first.

We achieve this in our \textbf{AnchorTree} framework by parametrizing the anchor weight \(\mu\) in the ANELBO training objective (\cref{eq:anelbo}) to position-dependent and of form
\begin{equation}
    \mu(l) = \omega (l) \cdot \eta (l),
\end{equation}
where \(\omega (l) \in \{0,1\}\) is an indicator denoting which positions are anchored, and \(\eta(l) > 0\) is a schedule monotonically decreasing with AST-depth with \(\eta(l') \geq \eta(l)\) if \(l' \succeq l\).
This yields a \textit{soft anchoring} scheme: we use \(\omega\) to select the anchor tokens, and then assign importance to them via \(\eta\).


Concretely, we let $\omega(l)=1$ for keyword and identifier tokens, which we visualize in~\cref{fig:anchor_strategies_gallery} (middle).
The choice of keywords is because this is the primary way programs denote \textit{syntactic structure}, e.g., \texttt{def} begins a function definition, while \texttt{if} signals an if-statement.
On the other hand, identifiers such as function names and local variables convey \textit{semantic intent}, especially in human-written code~\citep{hindle2016naturalness}.
Thus, the simultaneous anchoring of keywords and identifiers enables us to extract \textit{both} syntactic and semantic information from the AST, a data structure designed principally for syntactic analysis.


Next, we capture hierarchical ordering through the schedule \(\eta\) that monotonically decays with the AST depth of \(\msf{node}(l)\), which we visualize in~\cref{fig:anchor_strategies_gallery} (right).
As a concrete implementation in our experiments, we use an exponential depth-decay of the form
\begin{equation}
\label{eq:eta_exp_decay}
    \eta(l) = \gamma \cdot \exp (-\beta \cdot \mathrm{max}(\msf{depth}(l) - d_0, 0)),
\end{equation}
where $\gamma$ is tuned per anchoring strategy with a fixed $\beta=0.7$, \(\msf{depth}(l)\) is the AST depth of \(\msf{node}(l)\), and \(d_0 = 2\) is the AST depth at which decay begins.
Note that when \(\beta = 0\), we recover hard anchoring with \(\mu = \gamma\).
This parametrization of \(\eta\) ensures that tokens belonging to a parent node receive more weight than those in a child node, and are therefore prioritized by the anchor model.

Together, the combination of keyword and identifier anchor tokens with AST-informed depth weighting defines \textbf{AnchorTree}.
This hierarchical soft-anchoring framework for code diffusion serves as the basis for our \textbf{AnCoder} models, which we describe next.

\begin{table*}[!t]
\centering
\caption{\textbf{Evaluation on standard coding benchmarks.}
We report syntactic validity (Syntax (\%)) and functional correctness (Pass@1 (\%)) on HumanEval and MBPP at \(T = 2048\) diffusion steps.
AnchorTree substantially improves over hard anchoring (Keyword, Identifier) and null anchoring (\(\gamma = 0\)) baselines.
All 809M models use the two-stage anchor-denoiser architecture discussed in~\cref{sec:anchored-dlm}.
}
\label{tab:code_eval_benchmarks}
\resizebox{\textwidth}{!}{
\begin{tabular}{llccccc}
\toprule
& &  & \multicolumn{2}{c}{HumanEval} & \multicolumn{2}{c}{MBPP} \\
\cmidrule(lr){4-5} \cmidrule(lr){6-7}
Model & Size & Strategy & Syntax (\%) & Pass@1 (\%) & Syntax (\%) & Pass@1 (\%)\\
\midrule
\rowcolor{gray!20}
AR~\citep{hui2024qwen2} & 494M & --
    & -- & 30.50 & -- & 39.30
\\
\rowcolor{gray!20}
AR~\citep{hui2024qwen2} (re-eval) & 494M & --
    & 93.90 & 26.83 & 95.59 & 36.55
\\
\rowcolor{gray!20}
MDLM~\citep{sahoo2024simple} & 494M & --
    & 62.41 & 2.35 & 71.83 & 6.65
\\
MDLM~\citep{sahoo2024simple} & 809M & --
    & 65.58 & 3.29 & 74.46 & 6.34 \\
\rowcolor{orange!20}
AnCoder (ours) & 809M & Null
    & 72.24 & 4.80 & 79.11 & 8.64 \\
\rowcolor{orange!20}
AnCoder (ours) & 809M & Keyword
    & 73.05 & 4.57 & 78.47 & 8.81 \\
\rowcolor{orange!20}
AnCoder (ours) & 809M & Identifier
    & 72.65 & 4.73 & 79.25 & 8.95 \\
\rowcolor{orange!20}
{AnCoder} (ours) & {809M} & {AnchorTree}
    & \textbf{74.72} & \textbf{5.45} & \textbf{80.12} & \textbf{9.10} \\
\bottomrule
\end{tabular}
}
\end{table*}



\section{Experiments}
\label{sec:experiments}

In this section, we describe the application of our AnchorTree framework to train 809M-parameter AnCoder models and evaluate them against DLM baselines on coding benchmarks.

\textbf{Architecture.}
We initialize the AnCoder models with a two-staged anchor-denoiser architecture described in~\cref{sec:anchored-dlm}.
AnCoder models have 809M total parameters: the anchor network \(\theta_A\) has 494M parameters (24 transformer layers), initialized with the weights of Qwen2.5-Coder-0.5B~\citep{hui2024qwen2}; the denoiser network \(\theta_D\) has 315M parameters (12 layers), initialized with the first 12 layers of Qwen.
These parameter counts are because Qwen shares the 136M parameters of its word embedding with the LM head, and each of its 24 transformer layers has 14.9M parameters.
To enable diffusion language modeling, we convert Qwen's causal attention to a bidirectional one.
We truncate or pad all inputs to a context length of \(L = 1024\).




\textbf{Training.}
We use datasets from OpenCoder~\citep{huang2025opencoder} and split training into three stages: AR-to-DLM pretraining adaptation (197K steps), mid-training with a mixture of coding and mathematical reasoning tasks (171K steps), and finally SFT with a Python-focused dataset (50K steps).
We use AdamW~\citep{loshchilov2017decoupled} with: learning rate $10^{-4}$, batch size 512, cosine learning rate scheduler, and a warmup ratio of 0.01.
We use a cosine noise schedule~\citep{zhang2025cosine} for the diffusion noising process, where $\alpha_t = \cos^2\!\left(\tfrac{t}{T}\cdot\tfrac{\pi}{2}\right)$, which also informs our NELBO and ANELBO loss functions (see~\cref{eq:nelbo},~\cref{eq:anelbo}).
This setup was used to train AnCoder and the baseline methods.
We give additional dataset and training details in~\cref{sec:addn-exps}.

\textbf{Inference.}
We similarly use the cosine noise schedule for the backwards process.
We evaluate models for \(T \in \{256, 512, 1024, 2048\}\) backward denoising steps and a sampling temperature of 0.8.
For AnCoder implementations, we use a remask rate of 0.1, which we have found to improve performance; MDLM baselines use a remask rate of 0.0.
However, we do not incorporate the remask rate into the time-indexed weight \(\lambda_i\) of the DLM training loss (see~\cref{eq:nelbo}).

\textbf{Benchmarks.}
We evaluate on two standard code generation benchmarks:
\textit{HumanEval}~\citep{chen2021evaluating}, consisting of 164 Python programming problems, and
\textit{MBPP}~\citep{austin2021program}, consisting of 974 problems.
Each benchmark provides a natural-language prompt describing the task and a set of unit tests used for evaluation.
We report \emph{Pass@1}, defined as the probability that a generated solution passes all unit tests, as well as \emph{syntactic validity}, measuring whether the generated code parses successfully~\citep{chen2021evaluating}.
We generate 20 samples per prompt for HumanEval and 10 for MBPP.
Example prompts and unit tests are provided in Appendix~\ref{sec:addn-exps}.

\textbf{Baselines.}
Our primary baseline is MDLM~\citep{sahoo2024simple,md4}, which we re-train using the same backbone initialization and training recipe as AnCoder for a controlled comparison.
All 809M-models, including the 809M MDLM baseline, use the two-stage anchor-denoiser architecture described in~\cref{sec:anchored-dlm}.
We additionally compare against Qwen2.5-Coder-0.5B~\citep{hui2024qwen2}, the 494M AR model used to initialize AnCoder models.



\begin{table*}[!t]
\centering
\caption{\textbf{Scaling test-time compute on HumanEval.}
We report how the syntactic validity (Syntax (\%)) and functional correctness (Pass@1 (\%)) scale with the number of denoising steps \(T\).
In particular, using more denoising steps improves performance, with anchored models benefiting the most.
}
\label{tab:big_table_humaneval}
\setlength{\tabcolsep}{5pt}
\resizebox{\textwidth}{!}{
\begin{tabular}{llccc cc cc cc}
\toprule
& & & \multicolumn{2}{c}{$T=256$} & \multicolumn{2}{c}{$T=512$} & \multicolumn{2}{c}{$T=1024$} & \multicolumn{2}{c}{$T=2048$} \\
\cmidrule(lr){4-5}\cmidrule(lr){6-7}\cmidrule(lr){8-9}\cmidrule(lr){10-11}
Method & Size & Anchoring
& Syntax & Pass@1
& Syntax & Pass@1
& Syntax & Pass@1
& Syntax & Pass@1 \\
\midrule
\rowcolor{gray!20}
MDLM & 494M & --
& 57.80 & 2.32
& 61.68 & 2.35
& 61.71 & 2.71
& 62.41 & 2.35 \\ 



MDLM & 809M & --
& 61.83 & 3.23
& 62.53 & 3.02
& 64.33 & 2.84
& 65.58 & 3.29 \\ 


\rowcolor{orange!20}
AnCoder & 809M & Null
& 64.27 & 3.20
& 66.65 & 3.87
& 71.49 & 4.48
& 72.24 & 4.80 \\ 

\rowcolor{orange!20}
AnCoder & 809M & Keyword
& 62.07 & 2.96
& 66.25 & 3.17
& 69.57 & \textbf{4.88}
& 73.05 & 4.57 \\ 

\rowcolor{orange!20}
AnCoder & 809M & Identifier
& 63.17 & \textbf{3.38}
& 67.68 & 3.75
& 70.52 & 3.96
& 72.65 & 4.73 \\ 

\rowcolor{orange!20}
AnCoder & 809M & AnchorTree
& \textbf{64.48} & 3.14
& \textbf{68.75} & \textbf{4.15}
& \textbf{71.83} & 4.27
& \textbf{74.72} & \textbf{5.45} \\ 
\bottomrule
\end{tabular}
}
\end{table*}

\begin{table*}[!t]
\label{-1ex}
\centering
\caption{\textbf{Scaling test-time compute on MBPP}.
Similar to~\cref{tab:big_table_humaneval}, we report syntactic validity (Syntax (\%)) and functional correctness (Pass@1 (\%)) for different models at different numbers of denoising steps \(T\).
We similarly observe performance gain as \(T\) increases, with anchored models benefiting the most.
}
\label{tab:big_table_mbpp}
\setlength{\tabcolsep}{5pt}
\resizebox{\textwidth}{!}{
\begin{tabular}{llccc cc cc cc}
\toprule
& & & \multicolumn{2}{c}{$T=256$} & \multicolumn{2}{c}{$T=512$} & \multicolumn{2}{c}{$T=1024$} & \multicolumn{2}{c}{$T=2048$} \\
\cmidrule(lr){4-5}\cmidrule(lr){6-7}\cmidrule(lr){8-9}\cmidrule(lr){10-11}
Method & Size & Anchoring
& Syntax & Pass@1
& Syntax & Pass@1
& Syntax & Pass@1
& Syntax & Pass@1 \\
\midrule
\rowcolor{gray!20}
MDLM & 494M & --
& 67.65 & 6.29
& 69.73 & 6.45
& 71.27 & 6.74
& 71.83 & 6.65 \\ 



MDLM & 809M & --
& 69.76 & 5.95
& 71.99 & 6.61
& 72.73 & 6.62
& 74.46 & 6.34 \\ 


\rowcolor{orange!20}
AnCoder & 809M & Null
& 71.04 & \textbf{6.79}
& 75.42 & 7.00
& 78.52 & 7.81
& 79.11 & 8.64 \\ 

\rowcolor{orange!20}
AnCoder & 809M & Keyword
& 70.89 & 6.31
& 74.79 & 7.27
& 77.42 & 7.85
& 78.47 & 8.81 \\ 

\rowcolor{orange!20}
AnCoder & 809M & Identifier
& 72.08 & 6.60
& 75.54 & 7.10
& 77.78 & 8.30
& 79.25 & 8.95 \\ 

\rowcolor{orange!20}
AnCoder & 809M & AnchorTree 
& \textbf{72.97} & 6.74
& \textbf{76.62} & \textbf{7.39}
& \textbf{80.35} & \textbf{8.34}
& \textbf{80.12} & \textbf{9.10} \\ 
\bottomrule
\end{tabular}
}
\end{table*}

\subsection{Evaluation on Standard Benchmarks}
\label{sec:code_evals_benchmarks}

We show in~\cref{tab:code_eval_benchmarks} the performance of different models on the HumanEval and MBPP benchmarks.
In particular, AnCoder consistently outperforms the MDLM and anchoring baselines of the same scale, with the largest gains achieved by the AnchorTree anchoring scheme.


On HumanEval, AnCoder with AnchorTree improves syntactic validity from 65.58\% to 74.72\% (+9.14 absolute, +13.9\% relative) and Pass@1 from 3.29\% to 5.45\% (+2.16 absolute, +65.7\% relative) compared to MDLM at 809M.
Similar trends are observed on MBPP, where AnCoder improves syntactic validity from 74.46\% to 80.12\% (+5.66 absolute, +7.6\% relative) and pass@1 from 6.34\% to 9.10\% (+2.76 absolute, +43.5\% relative).
Notably, AnchorTree also outperforms other keyword- and identifier-based anchoring strategies, suggesting that soft anchoring is a better framework for code than hard anchoring ones.


However, all diffusion models fall short of the AR baseline in terms of syntactic validity and pass@1 on both benchmarks.
Nevertheless, AnCoder with AnchorTree shows the greatest improvement over the 809M MDLM baseline, suggesting that anchoring is a promising method for closing the well-documented AR-DLM performance gap~\citep{llada,li2025beyond}.


\subsection{Effect of Scaling Test-time Compute}
\label{sec:steps-humaneval}


A key advantage of DLMs is the ability to improve generation quality by taking more denoising steps.
We study how different baselines perform on HumanEval (\cref{tab:big_table_humaneval}) and MBPP (\cref{tab:big_table_mbpp}) for \(T \in \{256, 512, 1024, 2048\}\) denoising steps.

For MDLM, which does not implement remasking during denoising, we observe only modest improvements for both 494M and 809M models as \(T\) increases.
In fact, MDLM's performance can sometimes saturate, e.g., pass@1 on MBPP despite rising syntactic validity scores.
In contrast, AnCoder (Null) shows greater improvement with \(T\), suggesting the importance of a remask rate during denoising.
When anchoring is enabled (Keyword, Identifier, AnchorTree), this improvement is even greater, with AnchorTree attaining the strongest performance at \(T = 2048\) in syntactic validity and pass@1 on both benchmarks.

In summary, simply increasing \(T\) is \textit{insufficient} for high-quality DLM code generation.
Instead, effective test-time scaling requires improvements in both the sampling strategy (e.g., remask rate) and structural guidance, such as anchoring.

\begin{figure}[t]

\centering

\ifarxiv
    \includegraphics[width=0.6\linewidth]{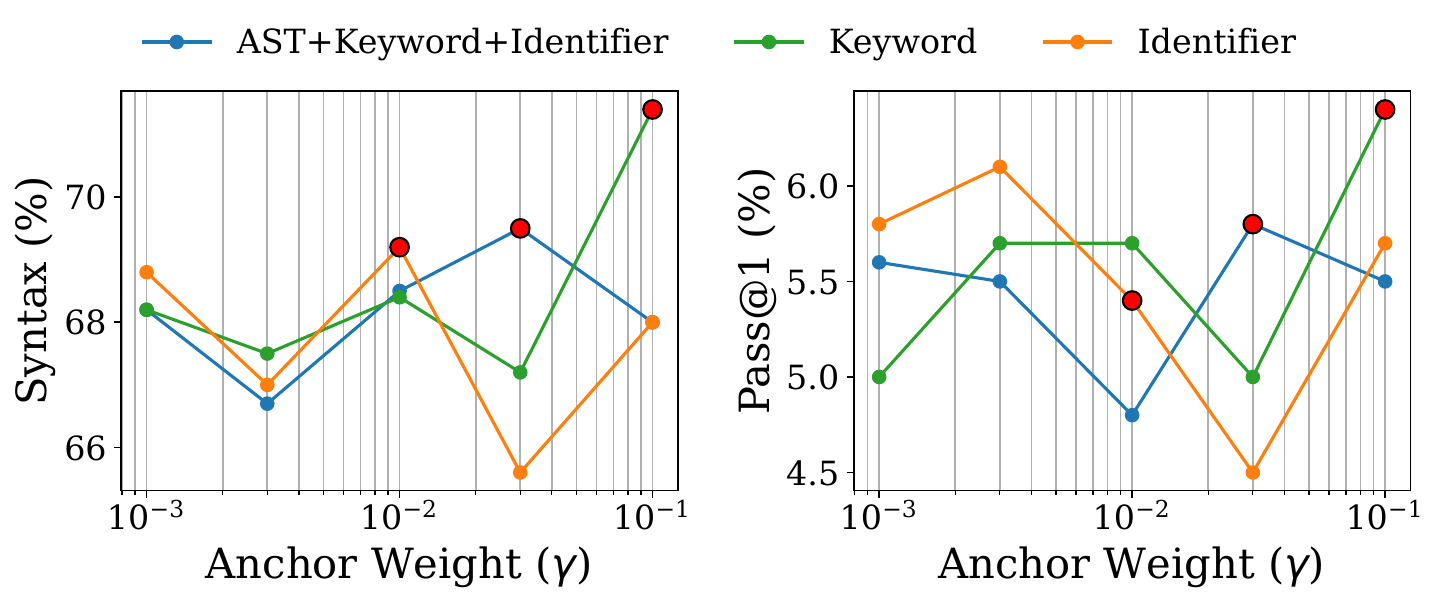}
\fi

\caption{
\textbf{Ablation on the anchor weight}.
Syntax (left) and Pass@1 (right) versus anchor weight $\gamma$ on HumanEval for three anchoring strategies.
Plots at $\gamma \in \{0.001,0.003,0.01,0.03,0.1\}$ on a log-$X$-axis.
}
\label{fig:anchor_ablation}
\end{figure}

\subsection{Ablation Study}
\label{sec:ablation}

We ablate the anchor supervision weight $\gamma$ for each anchoring strategy by training for two epochs on the OPC annealing dataset (12K steps) and evaluating on HumanEval.
Figure~\ref{fig:anchor_ablation} shows the resulting \emph{Syntax} (left) and \emph{Pass@1} (right) metrics as a function of the anchor weight.
The tabular results are reported in \cref{tab:anchor_ablation} (moved to the Appendix~\ref{sec:addn-ablation}); here we summarize the main trends and our chosen settings.
%
%
Based on the ablation study, we select $\gamma{=}0.03$ for AnchorTree, $\gamma{=}0.1$ for Keyword, and $\gamma{=}0.01$ for Identifier anchors. These configurations are highlighted in \cref{tab:anchor_ablation} and used in all benchmark evaluations.

\section{Discussion}
\label{sec:discussion}


\textbf{Anchored Architecture.}
There are several design choices in the two-stage anchor-denoiser network described in~\cref{sec:anchored-dlm}, particularly regarding how the anchor network passes information to the denoiser.
For instance, one may have the network carry over unmasked tokens, i.e., effectively treating all unmasked tokens as anchors for the denoiser.
Additionally, one may also implement a residual connection between the anchor and denoiser, in which case the anchor supervision is akin to a linear probe that steers the model.
There are also design choices in which model parameters to weight-tie, as the word embedding and LM head are often non-trivial (e.g., 136M for AnCoder).
In summary, the design space of anchored diffusion language models is still nascent, see~\citet{rout2025anchored} for further discussion.

\textbf{Anchor Token Selection and Weighting.}
While we have shown that keywords and identifiers are effective anchors for code generation, other choices may also be effective.
For instance, one might consider anchoring only on keywords related to control-flow, which would include ones such as \texttt{def}, \texttt{if}, and \texttt{for}, but exclude ones like \texttt{True}, \texttt{None}, and \texttt{import}.
Additionally, one may also consider anchoring only on specific identifiers, e.g., those present in function parameters and longer than \(2\) characters.
Additionally, while we have used a simple exponential depth-decay scheme for \(\eta\) (see~\cref{eq:eta_exp_decay}), other decay schedules may also be effective.

\textbf{Training and Optimization.}
It is known that training data selection and optimizer configuration significantly affect DLM performance~\citep{gong2025diffucoder,xie2025dream,chen2025coda,mercurydllm2025,li2025beyond}.
For instance, one might filter and curate higher-quality code datasets or employ multiple training stages beyond our current setup.
Indeed, we expect that such changes will also improve AnCoder, especially if one is to scale beyond the 1B-parameter regime, e.g., when initializing the DLM with larger pretrained AR models.
However, our work focuses on isolating and evaluating anchoring performance, which we achieve by building our implementation on a basic yet sufficient training and optimization pipeline.

\section{Conclusion}
\label{sec:conclusion}

We introduce AnchorTree, the first hierarchical soft anchoring framework tailored for code.
By anchoring on keywords and identifiers by AST hierarchy, AnchorTree captures both syntactic structure and semantic hierarchy while enabling performant coarse-to-fine diffusion denoising.
We validate this framework through a family of models that we call AnCoder, which empirically outperform DLM baselines on standard coding benchmarks.
Our findings suggest that using structural priors native to code, such as the AST, is an effective approach for parameter-efficient, high-quality code generation with discrete diffusion language models.



\section*{Acknowledgments}
This research has been supported by NSF Grants 2019844 and 2112471, the UT Austin Machine Learning Lab, and computing support on the Vista GPU Cluster through the Center for Generative AI (CGAI) and the Texas Advanced Computing Center (TACC) at UT Austin.


\bibliographystyle{plainnat}
\bibliography{references}

@article{md4,
  title={Simplified and generalized masked diffusion for discrete data},
  author={Shi, Jiaxin and Han, Kehang and Wang, Zhe and Doucet, Arnaud and Titsias, Michalis},
  journal={Advances in Neural Information Processing Systems},
  volume={37},
  pages={103131--103167},
  year={2024},
  url={https://arxiv.org/pdf/2406.04329}
}

@inproceedings{
remdm,
title={Remasking Discrete Diffusion Models with Inference-Time Scaling},
author={Guanghan Wang and Yair Schiff and Subham Sekhar Sahoo and Volodymyr Kuleshov},
booktitle={The Thirty-ninth Annual Conference on Neural Information Processing Systems},
year={2025},
url={https://openreview.net/forum?id=IJryQAOy0p}
}

@inproceedings{sohl2015deep,
  title={Deep unsupervised learning using nonequilibrium thermodynamics},
  author={Sohl-Dickstein, Jascha and Weiss, Eric and Maheswaranathan, Niru and Ganguli, Surya},
  booktitle={International conference on machine learning},
  pages={2256--2265},
  year={2015},
  organization={pmlr},
  url={https://proceedings.mlr.press/v37/sohl-dickstein15.pdf}
}

@techreport{geminidiffusion,
  author      = {{DeepMind}},
  title       = {Gemini Diffusion},
  institution = {DeepMind},
  year        = {2025},
  url         = {https://deepmind.google/models/gemini-diffusion/},
  note        = {Accessed: 2026-01-24}
}

@article{bavarian2022efficient,
  title={Efficient training of language models to fill in the middle},
  author={Bavarian, Mohammad and Jun, Heewoo and Tezak, Nikolas and Schulman, John and McLeavey, Christine and Tworek, Jerry and Chen, Mark},
  journal={arXiv preprint arXiv:2207.14255},
  year={2022},
  url={https://arxiv.org/pdf/2207.14255}
}

@inproceedings{ahmad2021unified,
  title={Unified pre-training for program understanding and generation},
  author={Ahmad, Wasi and Chakraborty, Saikat and Ray, Baishakhi and Chang, Kai-Wei},
  booktitle={Proceedings of the 2021 conference of the North American chapter of the association for computational linguistics: human language technologies},
  pages={2655--2668},
  year={2021},
  url={https://aclanthology.org/2021.naacl-main.211.pdf}
}

@article{alphacode,
  title={Competition-level code generation with alphacode},
  author={Li, Yujia and Choi, David and Chung, Junyoung and Kushman, Nate and Schrittwieser, Julian and Leblond, R{\'e}mi and Eccles, Tom and Keeling, James and Gimeno, Felix and Dal Lago, Agustin and others},
  journal={Science},
  volume={378},
  number={6624},
  pages={1092--1097},
  year={2022},
  publisher={American Association for the Advancement of Science},
  url={https://www.science.org/doi/epdf/10.1126/science.abq1158}
}

@inproceedings{
incoder,
title={InCoder: A Generative Model for Code Infilling and Synthesis},
author={Daniel Fried and Armen Aghajanyan and Jessy Lin and Sida Wang and Eric Wallace and Freda Shi and Ruiqi Zhong and Scott Yih and Luke Zettlemoyer and Mike Lewis},
booktitle={The Eleventh International Conference on Learning Representations },
year={2023},
url={https://openreview.net/forum?id=hQwb-lbM6EL}
}

@article{deepseek,
  title={Deepseek-r1: Incentivizing reasoning capability in llms via reinforcement learning},
  author={Guo, Daya and Yang, Dejian and Zhang, Haowei and Song, Junxiao and Zhang, Ruoyu and Xu, Runxin and Zhu, Qihao and Ma, Shirong and Wang, Peiyi and Bi, Xiao and others},
  journal={arXiv preprint arXiv:2501.12948},
  year={2025},
  url={https://arxiv.org/pdf/2501.12948}
}

@article{muller2014pystruct,
  title={PyStruct: learning structured prediction in python.},
  author={M{\"u}ller, Andreas C and Behnke, Sven},
  journal={J. Mach. Learn. Res.},
  volume={15},
  number={1},
  pages={2055--2060},
  year={2014},
  url={https://www.jmlr.org/papers/volume15/mueller14a/mueller14a.pdf}
}

@article{llada,
  title={Large language diffusion models},
  author={Nie, Shen and Zhu, Fengqi and You, Zebin and Zhang, Xiaolu and Ou, Jingyang and Hu, Jun and Zhou, Jun and Lin, Yankai and Wen, Ji-Rong and Li, Chongxuan},
  journal={arXiv preprint arXiv:2502.09992},
  year={2025}
}

@article{qwen3,
  title={Qwen3 technical report},
  author={Yang, An and Li, Anfeng and Yang, Baosong and Zhang, Beichen and Hui, Binyuan and Zheng, Bo and Yu, Bowen and Gao, Chang and Huang, Chengen and Lv, Chenxu and others},
  journal={arXiv preprint arXiv:2505.09388},
  year={2025},
  url={https://arxiv.org/pdf/2505.09388}
}

@techreport{gpt,
  author      = {{OpenAI}},
  title       = {OpenAI o3 and o4-mini System Card},
  institution = {OpenAI},
  year        = {2025},
  type        = {System Card},
  url         = {https://cdn.openai.com/pdf/2221c875-02dc-4789-800b-e7758f3722c1/o3-and-o4-mini-system-card.pdf},
  note        = {Accessed: 2026-01-24}
}

@article{claude,
  title={The claude 3 model family: Opus, sonnet, haiku},
  author={Anthropic, AI},
  journal={Claude-3 Model Card},
  volume={1},
  pages={1},
  year={2024},
  url={https://www-cdn.anthropic.com/de8ba9b01c9ab7cbabf5c33b80b7bbc618857627/Model_Card_Claude_3.pdf}
}

@article{gemini,
  title={Gemini 2.5: Pushing the frontier with advanced reasoning, multimodality, long context, and next generation agentic capabilities},
  author={Comanici, Gheorghe and Bieber, Eric and Schaekermann, Mike and Pasupat, Ice and Sachdeva, Noveen and Dhillon, Inderjit and Blistein, Marcel and Ram, Ori and Zhang, Dan and Rosen, Evan and others},
  journal={arXiv preprint arXiv:2507.06261},
  year={2025}
}

@inproceedings{rout2025anchored,
  title={Anchored Diffusion Language Model},
  author={Litu Rout and Constantine Caramanis and Sanjay Shakkottai},
  booktitle={The Thirty-ninth Annual Conference on Neural Information Processing Systems},
  year={2025},
  url={https://openreview.net/forum?id=E8adS5srds}
}

@article{rout2025test,
  title={Test-Time Anchoring for Discrete Diffusion Posterior Sampling},
  author={Rout, Litu and Lugmayr, Andreas and Jafarian, Yasamin and Varadharajan, Srivatsan and Caramanis, Constantine and Shakkottai, Sanjay and Kemelmacher-Shlizerman, Ira},
  journal={arXiv preprint arXiv:2510.02291},
  year={2025},
  url={https://arxiv.org/pdf/2510.02291}
}

@article{austin2021structured,
  title={Structured denoising diffusion models in discrete state-spaces},
  author={Austin, Jacob and Johnson, Daniel D and Ho, Jonathan and Tarlow, Daniel and Van Den Berg, Rianne},
  journal={Advances in neural information processing systems},
  volume={34},
  pages={17981--17993},
  year={2021}
}

@article{li2022diffusion,
  title={Diffusion-lm improves controllable text generation},
  author={Li, Xiang and Thickstun, John and Gulrajani, Ishaan and Liang, Percy S and Hashimoto, Tatsunori B},
  journal={Advances in neural information processing systems},
  volume={35},
  pages={4328--4343},
  year={2022}
}

@book{aho2007compilers,
  title={Compilers principles, techniques \& tools},
  author={Alfred, V Aho and Monica, S Lam and Jeffrey, D Ullman},
  year={2007},
  publisher={pearson Education}
}

@article{lou2023discrete,
  title={Discrete diffusion modeling by estimating the ratios of the data distribution},
  author={Lou, Aaron and Meng, Chenlin and Ermon, Stefano},
  journal={arXiv preprint arXiv:2310.16834},
  year={2023}
}

@article{ou2024your,
  title={Your absorbing discrete diffusion secretly models the conditional distributions of clean data},
  author={Ou, Jingyang and Nie, Shen and Xue, Kaiwen and Zhu, Fengqi and Sun, Jiacheng and Li, Zhenguo and Li, Chongxuan},
  journal={arXiv preprint arXiv:2406.03736},
  year={2024}
}

@article{shi2024simplified,
  title={Simplified and generalized masked diffusion for discrete data},
  author={Shi, Jiaxin and Han, Kehang and Wang, Zhe and Doucet, Arnaud and Titsias, Michalis},
  journal={Advances in neural information processing systems},
  volume={37},
  pages={103131--103167},
  year={2024}
}

@article{gong2025diffucoder,
  title={DiffuCoder: Understanding and Improving Masked Diffusion Models for Code Generation},
  author={Gong, Shansan and Zhang, Ruixiang and Zheng, Huangjie and Gu, Jiatao and Jaitly, Navdeep and Kong, Lingpeng and Zhang, Yizhe},
  journal={arXiv preprint arXiv:2506.20639},
  year={2025}
}

@article{sahoo2024simple,
  title={Simple and effective masked diffusion language models},
  author={Sahoo, Subham and Arriola, Marianne and Schiff, Yair and Gokaslan, Aaron and Marroquin, Edgar and Chiu, Justin and Rush, Alexander and Kuleshov, Volodymyr},
  journal={Advances in Neural Information Processing Systems},
  volume={37},
  pages={130136--130184},
  year={2024}
}

@article{xie2025dream,
  title={Dream-coder 7b: An open diffusion language model for code},
  author={Xie, Zhihui and Ye, Jiacheng and Zheng, Lin and Gao, Jiahui and Dong, Jingwei and Wu, Zirui and Zhao, Xueliang and Gong, Shansan and Jiang, Xin and Li, Zhenguo and others},
  journal={arXiv preprint arXiv:2509.01142},
  year={2025}
}

@article{chen2021evaluating,
  title={Evaluating large language models trained on code},
  author={Chen, Mark},
  journal={arXiv preprint arXiv:2107.03374},
  year={2021}
}

@article{austin2021program,
  title={Program synthesis with large language models},
  author={Austin, Jacob and Odena, Augustus and Nye, Maxwell and Bosma, Maarten and Michalewski, Henryk and Dohan, David and Jiang, Ellen and Cai, Carrie and Terry, Michael and Le, Quoc and others},
  journal={arXiv preprint arXiv:2108.07732},
  year={2021}
}

@article{chen2025coda,
  title={CoDA: Coding LM via Diffusion Adaptation},
  author={Chen, Haolin and Wang, Shiyu and Qin, Can and Pang, Bo and Liu, Zuxin and Qiu, Jielin and Zhang, Jianguo and Zhou, Yingbo and Chen, Zeyuan and Xu, Ran and others},
  journal={arXiv preprint arXiv:2510.03270},
  year={2025}
}

@article{mercurydllm2025,
  title = {Mercury: Ultra-Fast Language Models Based on Diffusion},
  author = {Khanna, Samar and Kharbanda, Siddhant and Li, Shufan and Varma, Harshit and Wang, Eric and Birnbaum, Sawyer and Luo, Ziyang and Miraoui, Yanis and Palrecha, Akash and Ermon, Stefano and Grover, Aditya and Kuleshov, Volodymyr},
  journal = {arXiv preprint arXiv:2506.17298},
  year = {2025},
}

@article{suresh2025dingo,
  title={DINGO: Constrained Inference for Diffusion LLMs},
  author={Suresh, Tarun and Banerjee, Debangshu and Ugare, Shubham and Misailovic, Sasa and Singh, Gagandeep},
  journal={arXiv preprint arXiv:2505.23061},
  year={2025}
}

@article{mundler2025constrained,
  title={Constrained decoding of diffusion llms with context-free grammars},
  author={M{\"u}ndler, Niels and Dekoninck, Jasper and Vechev, Martin},
  journal={arXiv preprint arXiv:2508.10111},
  year={2025},
  url={https://arxiv.org/pdf/2508.10111}
}

@article{hui2024qwen2,
  title={Qwen2. 5-coder technical report},
  author={Hui, Binyuan and Yang, Jian and Cui, Zeyu and Yang, Jiaxi and Liu, Dayiheng and Zhang, Lei and Liu, Tianyu and Zhang, Jiajun and Yu, Bowen and Lu, Keming and others},
  journal={arXiv preprint arXiv:2409.12186},
  year={2024}
}

@inproceedings{huang2025opencoder,
  title={Opencoder: The open cookbook for top-tier code large language models},
  author={Huang, Siming and Cheng, Tianhao and Liu, Jason Klein and Xu, Weidi and Hao, Jiaran and Song, Liuyihan and Xu, Yang and Yang, Jian and Liu, Jiaheng and Zhang, Chenchen and others},
  booktitle={Proceedings of the 63rd Annual Meeting of the Association for Computational Linguistics (Volume 1: Long Papers)},
  pages={33167--33193},
  year={2025}
}

@article{olausson2023self,
  title={Is self-repair a silver bullet for code generation?},
  author={Olausson, Theo X and Inala, Jeevana Priya and Wang, Chenglong and Gao, Jianfeng and Solar-Lezama, Armando},
  journal={arXiv preprint arXiv:2306.09896},
  year={2023}
}

@article{lam2025codecrash,
  title={CODECRASH: Stress Testing LLM Reasoning under Structural and Semantic Perturbations},
  author={Lam, Man Ho and Wang, Chaozheng and Huang, Jen-tse and Lyu, Michael R},
  journal={arXiv preprint arXiv:2504.14119},
  year={2025}
}

@inproceedings{sharifloo2025llms,
  title={Where Do LLMs Still Struggle? An In-Depth Analysis of Code Generation Benchmarks},
  author={Sharifloo, Amir Molzam and Heydari, Maedeh and Kazerooni, Parsa and Maninger, Daniel and Mezini, Mira},
  booktitle={2025 2nd IEEE/ACM International Conference on AI-powered Software (AIware)},
  pages={249--253},
  year={2025},
  organization={IEEE}
}

@book{scott2000programming,
  title={Programming language pragmatics},
  author={Scott, Michael},
  year={2000},
  publisher={Morgan Kaufmann}
}

@article{li2025beyond,
  title={Beyond autoregression: An empirical study of diffusion large language models for code generation},
  author={Li, Chengze and Zhang, Yitong and Li, Jia and Cai, Liyi and Li, Ge},
  journal={arXiv preprint arXiv:2509.11252},
  year={2025}
}

@book{cooper2022engineering,
  title={Engineering a compiler},
  author={Cooper, Keith D and Torczon, Linda},
  year={2022},
  publisher={Morgan Kaufmann}
}

@inproceedings{lattner2004llvm,
  title={LLVM: A compilation framework for lifelong program analysis \& transformation},
  author={Lattner, Chris and Adve, Vikram},
  booktitle={International symposium on code generation and optimization, 2004. CGO 2004.},
  pages={75--86},
  year={2004},
  organization={IEEE}
}

@article{casalnuovo2019studying,
  title={Studying the difference between natural and programming language corpora},
  author={Casalnuovo, Casey and Sagae, Kenji and Devanbu, Prem},
  journal={Empirical Software Engineering},
  volume={24},
  number={4},
  pages={1823--1868},
  year={2019},
  publisher={Springer}
}

@article{hindle2016naturalness,
  title={On the naturalness of software},
  author={Hindle, Abram and Barr, Earl T and Gabel, Mark and Su, Zhendong and Devanbu, Premkumar},
  journal={Communications of the ACM},
  volume={59},
  number={5},
  pages={122--131},
  year={2016},
  publisher={ACM New York, NY, USA}
}

@article{dong2025survey,
  title={A survey on code generation with llm-based agents},
  author={Dong, Yihong and Jiang, Xue and Qian, Jiaru and Wang, Tian and Zhang, Kechi and Jin, Zhi and Li, Ge},
  journal={arXiv preprint arXiv:2508.00083},
  year={2025}
}

@inproceedings{santa2025llm,
  title={Is LLM-Generated Code More Maintainable \& Reliable Than Human-Written Code?},
  author={Santa Molison, Alfred and Moraes, Marcia and Melo, Glaucia and Santos, Fabio and Assuncao, Wesley KG},
  booktitle={2025 ACM/IEEE International Symposium on Empirical Software Engineering and Measurement (ESEM)},
  pages={151--162},
  year={2025},
  organization={IEEE}
}

@inproceedings{siddiq2024quality,
  title={Quality assessment of chatgpt generated code and their use by developers},
  author={Siddiq, Mohammed Latif and Roney, Lindsay and Zhang, Jiahao and Santos, Joanna Cecilia Da Silva},
  booktitle={Proceedings of the 21st international conference on mining software repositories},
  pages={152--156},
  year={2024}
}

@article{abbassi2025unveiling,
  title={Unveiling inefficiencies in llm-generated code: Toward a comprehensive taxonomy},
  author={Abbassi, Altaf Allah and Da Silva, Leuson and Nikanjam, Amin and Khomh, Foutse},
  journal={arXiv preprint arXiv:2503.06327},
  year={2025}
}

@article{zeng2025treediff,
  title={TreeDiff: AST-Guided Code Generation with Diffusion LLMs},
  author={Zeng, Yiming and Cao, Jinghan and Li, Zexin and Chen, Yiming and Ren, Tao and Xiang, Dawei and Wu, Xidong and Gao, Shangqian and Yu, Tingting},
  journal={arXiv preprint arXiv:2508.01473},
  year={2025}
}

@article{zhang2025cosine,
  title={The cosine schedule is Fisher-Rao-optimal for masked discrete diffusion models},
  author={Zhang, Leo and Syed, Saifuddin},
  journal={arXiv preprint arXiv:2508.04884},
  year={2025}
}

@article{loshchilov2017decoupled,
  title={Decoupled weight decay regularization},
  author={Loshchilov, Ilya and Hutter, Frank},
  journal={arXiv preprint arXiv:1711.05101},
  year={2017}
}

@article{heinonen2023synthesizing,
  title={Synthesizing research on programmers’ mental models of programs, tasks and concepts—A systematic literature review},
  author={Heinonen, Ava and Lehtel{\"a}, Bettina and Hellas, Arto and Fagerholm, Fabian},
  journal={Information and Software Technology},
  volume={164},
  pages={107300},
  year={2023},
  publisher={Elsevier}
}

\newpage
\appendix
\onecolumn

\appendix
\section{Additional Experiments}
\label{sec:addn-exps}

This appendix provides supplementary material that supports the main results of the paper.
We include additional details on training data (Appendix~\ref{sec:addn-train-data}), benchmark descriptions (Appendix~\ref{sec:addn-benchmark}), hyperparameter ablations (Appendix~\ref{sec:addn-ablation}), and qualitative analyses of generated code (Appendix~\ref{sec:addn-qualitative-results}).

\subsection{Training and Fintuning Dataset}
\label{sec:addn-train-data}
\subsubsection{Stage 1: Pretraining}
\label{sec:stage1_data}

Stage 1 pretraining is used to adapt an autoregressive code model to the diffusion language modeling paradigm.
For this stage, we train on the \texttt{opc-fineweb-code} corpus from OpenCoder~\citep{huang2025opencoder}, which consists of 101M code examples drawn from web-scale programming data, with an average of 490 tokens per example (approximately 49.49B tokens in total).

This corpus provides broad coverage of real-world programming patterns, libraries, and coding styles, making it well-suited for initializing diffusion-based code generation.
During this stage, the model learns to denoise masked code sequences while retaining general-purpose code modeling capabilities inherited from autoregressive pretraining.
We train for a single epoch (197K optimization steps), which we find sufficient for stable AR-to-DLM adaptation.

\subsubsection{Stage: Mid-training}
\label{sec:mid-training_data}

Stage 2 mid-training is designed to bridge large-scale code pretraining and task-specific supervised fine-tuning by exposing the model to a diverse mixture of structured, synthetic, and web-scale data.
This stage is particularly important for diffusion-based models, as it encourages robust denoising behavior across a wide range of program structures and semantic patterns before task-specific specialization.

As summarized in Table~\ref{tab:mid-training_data}, the mid-training corpus consists of 29.49M examples drawn from the OpenCoder annealing datasets, comprising approximately 10.33B tokens in total.
The data are split into 99\% training and 1\% validation sets and are mixed uniformly during training.
Each component serves a distinct purpose:

\textbf{Algorithmic Code.}
This subset contains algorithmic programming problems with well-defined control flow and data dependencies.
It emphasizes precise reasoning, recursion, and structured program logic, which are essential for learning long-range dependencies and hierarchical program structure.

\textbf{Synthetic Code Snippets.}
The synthetic code subset consists of automatically generated code fragments covering a wide range of programming constructs.
These examples increase structural diversity and expose the model to varied syntactic patterns, helping prevent overfitting to narrow code distributions.

\textbf{Synthetic Question Answering.}
The synthetic QA data pairs natural-language questions with code-related answers.
This subset improves alignment between textual specifications and code generation, which is particularly relevant for benchmark-style prompts such as HumanEval and MBPP.

\textbf{FineWeb Math.}
The math corpus provides structured mathematical reasoning problems, encouraging the model to handle symbolic expressions, arithmetic logic, and step-by-step computation.
These capabilities are complementary to code generation and support precise denoising of structured sequences.

\textbf{FineWeb Code (sampled).}
The sampled FineWeb code data introduces real-world, web-scale programming content, capturing practical coding styles and idioms that are underrepresented in purely synthetic or algorithmic datasets.

Overall, this mixture is intended to balance structural rigor, semantic diversity, and real-world variability.
By mid-training on this heterogeneous corpus, AnCoder learns to denoise across a broad range of program structures, providing a strong foundation for the subsequent supervised fine-tuning stage.
We train for three epochs, for a total of approximately 171K optimization steps.

\begin{table}[t]
\centering
\caption{Datasets used during Stage 2 mid-training of AnCoder. We report the number of examples and the average number of tokens per example for each component of the OpenCoder annealing corpus (truncated to 1024 tokens).}
\label{tab:mid-training_data}
\setlength{\tabcolsep}{6pt}
\begin{tabular}{lccc}
\toprule
Dataset & Samples & Tokens & Domain \\
\midrule
Algorithmic Code & 5.32M & 1.01B & Algorithmic programs \\ 
Synthetic Code Snippets & 3.08M & 1.21B & Generated code \\ 
Synthetic QA & 10.60M & 3.74B & Code-related QA \\ 
FineWeb Math & 5.24M & 2.44B & Mathematical reasoning \\ 
FineWeb Code (sampled) & 5.24M & 1.93B & Web-scale code \\ 
\midrule
\textbf{Total} & \textbf{29.49M} & \textbf{10.33B} & -- \\
\bottomrule
\end{tabular}
\end{table}


\subsubsection{Stage 3: Supervised Fine-Tuning}
\label{sec:sft_data}

Stage 3 supervised fine-tuning (SFT) specializes the diffusion model for downstream Python code generation tasks.
We use the Python subset of the \texttt{opc-annealing-corpus}, including algorithmic code, synthetic code snippets, and synthetic question-answering data.
This subset contains 3.35M examples totaling approximately 1.1B tokens.

SFT examples are formatted as instruction-style prompts paired with target code solutions, aligning training with the evaluation format used in benchmarks such as HumanEval and MBPP.
We fine-tune for 8 epochs using a 95\%-5\% training-validation split, resulting in approximately 50K optimization steps.
This stage encourages precise alignment between natural-language specifications and executable code, complementing the more structure-focused objectives of the pretraining and midtraining stages.

\subsection{Evaluation Benchmarks}
\label{sec:addn-benchmark}

We evaluate AnCoder on two standard code generation benchmarks that assess both syntactic correctness and functional behavior of generated programs.
Below, we briefly describe these benchmarks, the evaluation protocol, and dataset usage.

\textbf{HumanEval.}
HumanEval~\citep{chen2021evaluating} is a benchmark of 164 Python programming problems designed to evaluate functional correctness of code generated from natural-language specifications.
Each problem consists of a function signature accompanied by a docstring that describes the intended behavior, along with a set of hidden unit tests used for evaluation.
A generated solution is considered correct if it passes all unit tests without raising runtime or syntax errors.
HumanEval emphasizes algorithmic reasoning, control flow, and correct handling of edge cases.

\textbf{MBPP.}
MBPP (Mostly Basic Python Programs)~\citep{austin2021program} consists of 974 Python programming tasks covering basic algorithmic and procedural concepts.
Each task provides a short natural-language description and multiple unit tests.
Compared to HumanEval, MBPP focuses on simpler programs but offers greater diversity in problem types, making it a useful complementary benchmark for assessing robustness and generalization.

\textbf{Evaluation Metrics.}
We report two primary metrics:
\emph{syntactic validity}, defined as the fraction of generated programs that successfully parse via the Python \texttt{ast} module's \texttt{ast.parse} without syntax errors, and
\emph{Pass@1}, defined as the probability that a single generated sample passes all provided unit tests.
Syntax errors are treated as failures under Pass@1, since such programs cannot execute successfully.

For each benchmark, we generate multiple samples per problem and estimate Pass@1 following standard practice, which is the average number of generations that pass the unit test.
Unless otherwise specified, we generate 20 samples per problem for HumanEval and 10 samples per problem for MBPP.

\textbf{Licensing and Usage.}
All datasets used in this work are publicly available and licensed for research use.
HumanEval is released under the MIT License.
MBPP is released under the Apache 2.0 License.
We use both datasets solely for non-commercial research and evaluation purposes, in accordance with their respective licenses.

\subsection{Ablation Study of Anchor Weight}
\label{sec:addn-ablation}

We conduct an ablation over the anchor supervision weight $\gamma$ for each anchoring strategy by training AnCoder for two epochs on the OPC annealing dataset (12K steps) and evaluating on HumanEval.
Table~\ref{tab:anchor_ablation} reports the resulting syntactic validity and Pass@1 for all tested values of $\gamma$, while Figure~\ref{fig:anchor_ablation} in the main paper visualizes the same trends.

For the AnchorTree strategy, performance exhibits a clear unimodal behavior.
Both syntactic validity and Pass@1 improve as $\gamma$ increases from $0.001$ to $0.03$, after which performance degrades slightly.
This suggests that moderate anchor supervision is sufficient to guide early denoising, while overly strong supervision can over-regularize the intermediate anchor prediction.

For Keyword anchoring, larger values of $\gamma$ are beneficial.
We observe the highest syntactic validity and Pass@1 at $\gamma=0.1$, indicating that when anchors are defined at the token level, stronger supervision is required to consistently influence the denoising process.

For Identifier anchoring, the two metrics peak at different values of $\gamma$.
Syntactic validity is maximized at $\gamma=0.01$, whereas Pass@1 attains its maximum at $\gamma=0.003$.
This highlights a trade-off between enforcing identifier structure and preserving flexibility for downstream functional correctness.

Overall, these results confirm that anchor supervision must be carefully calibrated.
Very small values of $\gamma$ underutilize the anchor signal, while overly large values can impede later denoising steps.
Based on this ablation, we select $\gamma=0.03$ for AnchorTree, $\gamma=0.1$ for Keyword, and $\gamma=0.01$ for Identifier anchoring strategies, and use these settings in all subsequent experiments.

\begin{table*}[!t]
    \centering
    \caption{Anchor weight $\gamma$ ablation on HumanEval after two epochs of training on the OPC annealing dataset (12K steps). We report syntactic validity and pass rate. The selected configuration for each anchor strategy is highlighted.}
    \label{tab:anchor_ablation}
    \begin{tabular}{llcc}
        \toprule
        Anchor Strategy & Anchor Weight ($\gamma$) & Syntax (\%) & Pass@1 (\%) \\
        \midrule
        \multirow{5}{*}{AnchorTree}
        & 0.001 & 68.2 & 5.6 \\
        & 0.003 & 66.7 & 5.5 \\
        & 0.01  & 68.5 & 4.8 \\
        & {\color{orange}0.03} & \textbf{69.5} & \textbf{5.8} \\
        & 0.1   & 68.0 & 5.5 \\
        \midrule
        \multirow{5}{*}{Keyword}
        & 0.001 & 68.2 & 5.0 \\
        & 0.003 & 67.5 & 5.7 \\
        & 0.01  & 68.4 & 5.7 \\
        & 0.03  & 67.2 & 5.0 \\
        & {\color{orange}0.1}  & \textbf{71.4} & \textbf{6.4} \\
        \midrule
        \multirow{5}{*}{Identifier}
        & 0.001 & 68.8 & 5.8 \\
        & 0.003 & 67.0 & \textbf{6.1} \\
        & {\color{orange}0.01} & \textbf{69.2} & 5.4 \\
        & 0.03  & 65.6 & 4.5 \\
        & 0.1   & 68.0 & 5.7 \\
        \bottomrule
    \end{tabular}
\end{table*}





\subsection{Qualitative Analysis of Generated Code}
\label{sec:addn-qualitative-results}

We qualitatively analyze code generated by standard diffusion and AnCoder variants to better understand the effects of hierarchical anchoring on program structure.
All models considered in this analysis have the same parameter count (809M) and are evaluated under identical inference settings.
Unless otherwise specified, all examples are generated with $T=2048$ denoising steps.
\begin{taskbox}
\textbf{Function:} \texttt{has\_close\_elements}
\begin{lstlisting}[style=pythonstyle]
def has_close_elements(numbers: List[float], threshold: float) -> bool:
    """ Check if in given list of numbers, are any two numbers closer to each other than given threshold.
    >>> has_close_elements([1.0, 2.0, 3.0], 0.5)
    False
    >>> has_close_elements([1.0, 2.8, 3.0, 4.0, 5.0, 2.0], 0.3)
    True
    """
\end{lstlisting}
\end{taskbox}

\subsubsection{HumanEval Qualitative Examples}
\label{sec:humaneval_qualitative}
We examine a representative example from the HumanEval benchmark to illustrate how different anchoring strategies influence the structural and logical properties of generated code.
The task is as follows: \textit{Given a list of numbers and a threshold, determine whether any two numbers in the list differ by less than the given threshold.}

\textbf{MDLM} produces syntactically valid code but fails to implement the intended logic.
The generated code introduces unnecessary control flow and incorrect numerical reasoning, resulting in logically invalid behavior despite parsing successfully.

\begin{codebox}
\begin{lstlisting}[style=pythonstyle]
def has_close_elements(numbers, threshold):
    from collections import Counter
    epsilon = float('inf')
    counter = Counter(numbers)
    while epsilon > 0:
        epsilon = 0.0001
        for num, count in counter.items():
            if num + epsilon in counter and abs(num - epsilon) < threshold:
                return True
        epsilon -= 1
    return False
\end{lstlisting}
\end{codebox}

\textbf{AnCoder (Null).}
With the two-stage architecture but no anchor supervision, AnCoder produces simpler code, but still fails to capture the core algorithmic structure.
The implementation mixes index-based reasoning with threshold comparisons in a way that does not correspond to the problem specification.

\begin{codebox}
\begin{lstlisting}[style=pythonstyle]
def has_close_elements(numbers, threshold):
    for i in enumerate(sorted(numbers), 1):
        if abs(i - numbers[i - 1]) == abs(i - threshold):
            return True
    return False
\end{lstlisting}
\end{codebox}

\textbf{AnCoder (Keyword).}
Keyword anchoring encourages the model to introduce appropriate control flow constructs, but the resulting logic remains overly coarse.
The model reduces the problem to global extrema comparisons, which is insufficient for detecting close neighboring values.

\begin{codebox}
\begin{lstlisting}[style=pythonstyle]
def has_close_elements(numbers, threshold):
    min_num = min(numbers)
    max_num = max(numbers)
    return max_num - min_num <= threshold or min_num - max_num <= threshold
\end{lstlisting}
\end{codebox}

\textbf{AnCoder (Identifier).}
Identifier anchoring improves variable usage consistency but still struggles to recover the correct relational structure between elements.
The model introduces compact expressions, but these do not faithfully implement the required pairwise comparison.

\begin{codebox}
\begin{lstlisting}[style=pythonstyle]
def has_close_elements(numbers, threshold):
    return min(num + 1 - numbers.index(num) for num in numbers) <= threshold
\end{lstlisting}
\end{codebox}

\textbf{AnCoder (AnchorTree).}
AnchorTree produces a correct and concise implementation.
The model recovers the intended algorithmic structure by sorting the input and comparing adjacent elements, closely matching human-written code.

\begin{codebox}
\begin{lstlisting}[style=pythonstyle]
def has_close_elements(numbers, threshold):
    numbers = sorted(numbers)
    for i in range(len(numbers) - 1):
        if abs(numbers[i] - numbers[i + 1]) < threshold:
            return True
    return False
\end{lstlisting}
\end{codebox}

\textbf{Summary.}
These examples illustrate that while standard diffusion and weak anchoring strategies can produce syntactically valid code, they often fail to recover the hierarchical structure required for correct execution.
In contrast, AnchorTree empowers AnCoder to identify and reveal structurally important tokens, early in the denoising process. Thus, AnCoder results in correct control flow and variable dependencies, often matching standard human-written solutions.

\subsubsection{MBPP Qualitative Examples}
\label{sec:mbpp_qualitative}
We analyze a representative example from the MBPP benchmark to further assess how anchoring influences structured code generation.
The task is as follows: \textit{Given a cost matrix and two positions, write a function to determine the minimum cost path between the two positions.}

\begin{taskbox}
\textbf{Function:} \texttt{min\_Cost}
\begin{lstlisting}[style=pythonstyle]
    Write a function to find the minimum cost path to reach (m, n) from (0, 0) for the given cost matrix cost[][] and a position (m, n) in cost[][].

    assert min_cost([[1, 2, 3], [4, 8, 2], [1, 5, 3]], 2, 2) == 8
    assert min_cost([[2, 3, 4], [5, 9, 3], [2, 6, 4]], 2, 2) == 12
    assert min_cost([[3, 4, 5], [6, 10, 4], [3, 7, 5]], 2, 2) == 16
    ```python
\end{lstlisting}
\end{taskbox}

\textbf{MDLM} produces syntactically valid code but fails to recover a coherent algorithmic structure.
The generated implementation contains incomplete control flow and undefined state transitions, leading to incorrect behavior.

\begin{codebox}
\begin{lstlisting}[style=pythonstyle]
def min_cost(cost, m, n):
    q = list([(0, (), 0)])
    position = [(0, 0)]
    for (x, pos), cost in q[-1]:
        if pos == (m, n):
            return cost
        q.append((x + cost, pos))
\end{lstlisting}
\end{codebox}

\textbf{AnCoder (Null).}
Without anchor supervision, AnCoder introduces a plausible high-level strategy using a priority queue, but the implementation suffers from inconsistent indexing and incorrect termination conditions.

\begin{codebox}
\begin{lstlisting}[style=pythonstyle]
import heapq
def min_cost(cost, m, n):
    q = []
    visited = [[False] * m for _ in range(n)]
    heapq.heappush(q, (cost[0][0], 0, 0))
    while q:
        c, x, y = heapq.heappop(q)
        if x == m and y == n:
            return c
        for dx, dy in [(1,0), (-1,0), (0,1), (0,-1)]:
            nx, ny = x + dx, y + dy
            if 0 <= nx < m and 0 <= ny < n:
                heapq.heappush(q, (c + cost[nx][ny], nx, ny))
\end{lstlisting}
\end{codebox}

\textbf{AnCoder (Keyword).}
Keyword anchoring encourages the use of explicit control flow constructs, but the model still fails to implement correct state transitions and stopping criteria.

\begin{codebox}
\begin{lstlisting}[style=pythonstyle]
from collections import deque
def min_path(cost, m, n):
    q = deque([(0, 0)])
    visited = set()
    while q:
        x, y = q.popleft()
        if (x, y) == (m, n):
            return 1
        for dx, dy in [(-1,0), (0,1), (1,0), (0,-1)]:
            q.append((x + dx, y + dy))
    return 0
\end{lstlisting}
\end{codebox}

\textbf{AnCoder (Identifier).}
Identifier anchoring improves variable consistency and reuse, but the generated code still fails to recover the correct recurrence relation and introduces unnecessary complexity.

\begin{codebox}
\begin{lstlisting}[style=pythonstyle]
def find_min_cost(cost, m, n):
    n = len(cost)
    costs = [[float('inf')] * n for _ in range(n)]
    def dfs(i, j):
        if costs[i][j] != float('inf'):
            return costs[i][j]
        costs[i][j] = cost[i][j] + min(dfs(i-1,j), dfs(i,j-1))
        return costs[i][j]
    return dfs(m, n)
\end{lstlisting}
\end{codebox}

\textbf{AnCoder (AnchorTree).}
AnchorTree enables AnCoder to produce a correct and concise solution with a clear recursive structure and appropriate base cases.
Interestingly, the generated code closely matches standard dynamic programming formulation for finding a minimum cost path between two positions.

\begin{codebox}
\begin{lstlisting}[style=pythonstyle]
def min_cost(cost, m, n):
    if m < 0 or n < 0:
        return float('inf')
    if m == 0 and n == 0:
        return cost[0][0]
    return cost[m][n] + min(
        min_cost(cost, m - 1, n),
        min_cost(cost, m, n - 1),
        min_cost(cost, m - 1, n - 1),
    )
\end{lstlisting}
\end{codebox}

\textbf{Summary.}
These examples mirror the trends observed on HumanEval.
The standard masked diffusion model and weak anchoring strategies often generate syntactically valid but structurally incorrect code, whereas AnchorTree enables AnCoder to recover hierarchical algorithmic structure, resulting in correct and interpretable solutions.




\subsection{Limitations}
\label{sec:failure_modes}

While AnchorTree consistently improves the syntactic validity and functional correctness of DLMs, there exist several limitations.
First, the development of anchored diffusion models is still nascent, and there exists a large search space of architecture design that could better utilize the AnchorTree framework.
Furthermore, AnchorTree is explicitly focused on code hierarchies extracted from the abstract syntax tree, which may limit its ability to reason about dynamic execution behavior.
Fortunately, many classical algorithms for analyzing such behavior rely on the construction of hierarchical data structures.
We thus expect that AnchorTree is also extendable to these settings.

\end{document}